\documentclass[sigconf]{acmart}
\AtBeginDocument{%
  }

\setcopyright{acmlicensed}
\copyrightyear{2025}
\acmYear{2025}
\acmDOI{XXXXXXX.XXXXXXX}

\usepackage{booktabs}
\usepackage{multirow}
\usepackage{array}
\usepackage{caption}
\usepackage{stfloats}
\begin{document}

\title{LightSearcher: Efficient DeepSearch via Experiential Memory}

\author{Hengzhi Lan$^{1,2}$, Yue Yu$^{1}$, Li Qian$^{3}$, Li Peng$^{3}$, Jie Wu$^{3}$, Wei Liu$^{3}$, Jian Luan$^{3}$, Ting Bai$^{1,2\ast}$}

\affiliation{%
  \vspace{1em}  
  \institution{$^{1}$ BaiJia AI Team, Beijing, China}
  \institution{$^{2}$ Beijing University of Posts and Telecommunications, Beijing, China}
  \institution{$^{3}$ Researcher}
  \country{}
}

\thanks{$^\ast$Corresponding author: Ting Bai}

\renewcommand{\shortauthors}{Trovato et al.}

\begin{abstract}
DeepSearch paradigms have become a core enabler for deep reasoning models, allowing them to invoke external search tools to access up-to-date, domain-specific knowledge beyond parametric boundaries, thereby enhancing the depth and factual reliability of reasoning.
Building upon this foundation, recent advances in reinforcement learning (RL) have further empowered models to autonomously and strategically control search tool usage, optimizing when and how to query external knowledge sources. Yet, these RL-driven DeepSearch systems often reveal a see-saw trade-off between accuracy and efficiency—frequent tool invocations can improve factual correctness but lead to unnecessary computational overhead and diminished efficiency.
To address this challenge, we propose \textbf{LightSearcher}, an efficient RL framework that incorporates textual experiential memory by learning contrastive reasoning trajectories to generate interpretable summaries of successful reasoning patterns. In addition, it employs an adaptive reward shaping mechanism that penalizes redundant tool calls only in correct-answer scenarios.
This design effectively balances the inherent accuracy–efficiency trade-off in DeepSearch paradigms. Experiments on four multi-hop QA benchmarks show that LightSearcher maintains accuracy comparable to SOTA baseline ReSearch, while reducing search tool invocations by \textbf{39.6\%}, inference time by \textbf{48.6\%}, and token consumption by \textbf{21.2\%}, demonstrating its superior efficiency.

\end{abstract}





\maketitle

\section{Introduction}

Deep reasoning models have showcased remarkable capabilities across a wide range of tasks~\cite{measuring,deepseek-r1}, yet they are inherently constrained by their parametric knowledge—struggling to access up-to-date information, domain-specific insights, or fact-intensive details critical for comprehensive and reliable responses~\cite{medical-research,llm-in-finance}. As a core enabler for overcoming this limitation, DeepSearch paradigms have become indispensable in advancing large reasoning models’ performance: by enabling models to invoke external search tools, they break through parametric boundaries to integrate external knowledge, thereby substantially enhancing the depth and factual reliability of reasoning.

\begin{figure}
  \centering
  \includegraphics[width=0.43\textwidth]{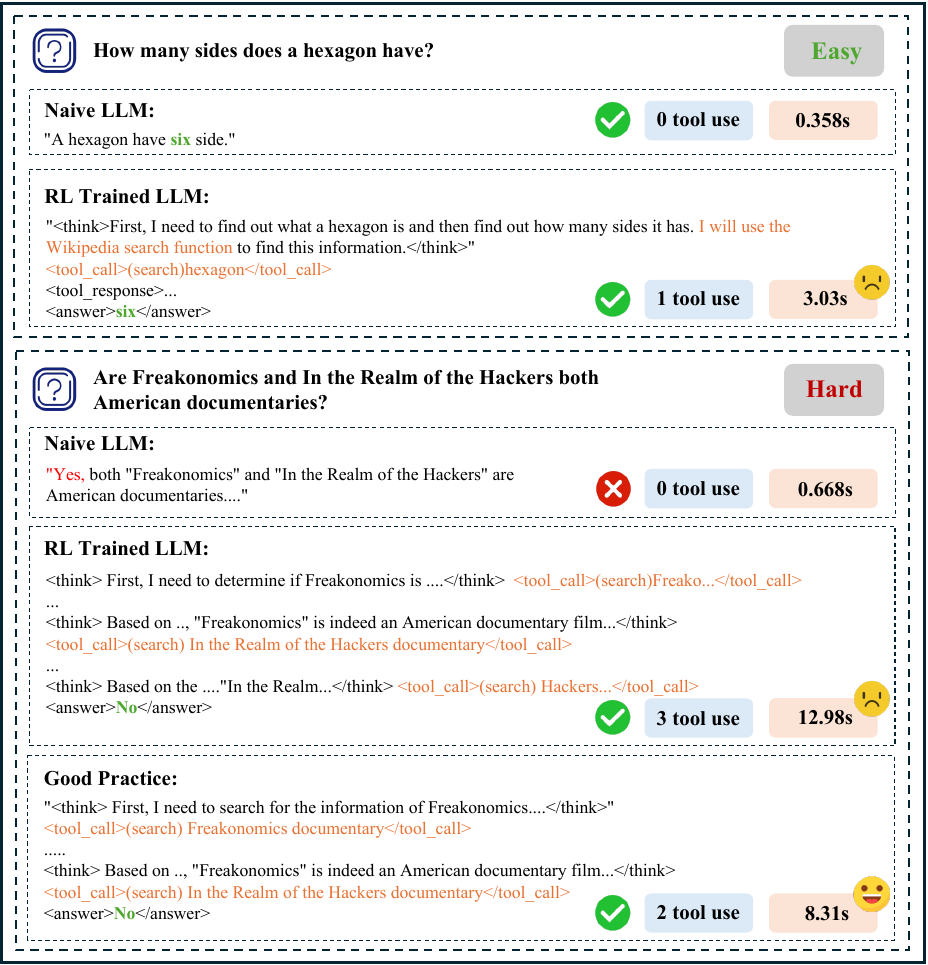}
  \caption{Illustration of Excessive Search Tool Usage in existing DeepSearch systems across both Easy and Hard queries, leading to degraded efficiency in deep reasoning models.
  }
  \label{fig:problem}
\end{figure}

To fully leverage the potential of DeepSearch, mainstream methodologies have explored Retrieval-Augmented Generation (RAG) techniques for integrating externally retrieved information into the reasoning pipeline~\cite{rag,rag-survey,kg-retriever,iter-retgen,emotional-rag,memoryos}. Early paradigms relied on supervised learning with manually annotated reasoning chains~\cite{ircot,iter-retgen} to guide tool invocation and retrieval. However, these suffer from high annotation costs and poor generalization, as manually crafted chains cannot adapt to the diversity of real-world queries~\cite{research}.
Recent advances in reinforcement learning (RL) have mitigated these limitations by enabling models to autonomously and strategically regulate search tool utilization, while simultaneously learning optimal policies for determining when and how to query external knowledge sources~\cite{research,search-r1}. Representative methods such as ReSearch~\cite{research} and Search-R1~\cite{search-r1} have yielded substantial performance gains on multi-hop question answering benchmarks~\cite{nq, hotpotqa, musique, 2wiki}.

However, these RL-driven DeepSearch approaches face a critical dilemma: a see-saw trade-off between accuracy and efficiency. As illustrated in Fig.~\ref{fig:problem}, models often exhibit excessive and indiscriminate search tool calls—resorting to retrieval even for queries that can be adequately answered using their intrinsic parametric knowledge. This over-reliance stems from the limitations of standard RL reward function designs, which primarily prioritize answer correctness. To maximize accuracy, models tend to increase tool invocation frequency, leading to unnecessary computational overhead, elevated token consumption, and diminished reasoning efficiency.
While recent studies have attempted to mitigate this issue with simple efficiency penalties~\cite{otc-po}, such approaches often result in performance degradation, as scalar reward optimization fails to fundamentally balance the dual objectives of accuracy and efficiency.

To address this unmet challenge, we propose \textbf{LightSearcher}, an efficient RL framework tailored for DeepSearch paradigms. LightSearcher integrates textual experiential memory by learning contrastive reasoning trajectories, distilling interpretable summaries of successful tool-invocation and reasoning patterns. Furthermore, it incorporates an adaptive reward shaping mechanism that penalizes redundant tool calls exclusively in correct-answer scenarios—avoiding efficiency sacrifices when accuracy is not yet achieved. By fusing experiential memory guidance with adaptive reward optimization, LightSearcher effectively resolves the inherent accuracy–efficiency trade-off in DeepSearch. Our key contributions are summarized as follows:
\begin{itemize}
\item We propose LightSearcher, an efficient RL framework tailored for DeepSearch, which integrates contrastive experiential memory to deliver explicit and interpretable guidance for the optimization of autonomous search tool invocation.
\item We design a novel adaptive reward shaping mechanism that dynamically balances accuracy and efficiency, penalizing redundant tool usage only when answers are correct.
\item Comprehensive experiments on four multi-hop QA benchmarks demonstrate that LightSearcher reduces search tool invocations by 39.6\% while maintaining comparable accuracy to the state-of-the-art baseline ReSearch, verifying its superiority in model efficiency.
\end{itemize}

\section{Related Work}
This section surveys key advancements in the DeepSearch area, emphasizing the integration of external knowledge with LLMs, efficiency challenges in tool invocation, and experiential memory in self-evolution LLMs.

\subsection{DeepSearch}
DeepSearch is an advanced reasoning paradigm integrating autonomous search mechanisms and iterative inference, empowering models to proactively retrieve, integrate, and validate external knowledge for complex tasks—particularly multi-hop reasoning, knowledge-intensive QA, and decision-making requiring progressive information accumulation~\cite{rag,rag-survey,ircot}. Synergizing external knowledge retrieval and internal reasoning, it addresses inherent limitations of large language models (LLMs), including outdated knowledge, poor logical consistency, and inability to handle complex reasoning chain dependencies.

Current DeepSearch methods fall into two main paradigms. The first uses prompting or fine-tuning to enable iterative search and generation: IRCoT~\cite{ircot} applies chain-of-thought reasoning to build intermediate logical chains for multi-hop retrieval; Iter-RetGen~\cite{iter-retgen} refines queries from intermediate answers to address dependencies; and Self-RAG~\cite{self-rag} incorporates self-reflection for autonomous retrieval decisions and quality assessment. Although effective in coordinating retrieval and generation, these methods depend on heuristic prompts and extensive annotated trajectories, hindering scalability. Inspired by recent reasoning training techniques, the second category employs reinforcement learning to fuse reasoning and retrieval: ReSearch~\cite{research}, Search-R1~\cite{search-r1}, and R1-searcher~\cite{r1-searcher}. These leverage RL algorithms like PPO~\cite{ppo}, GRPO~\cite{grpo}, and REINFORCE variants~\cite{reinforce,reinforce-plus-plus}, markedly improving models' ability to integrate external knowledge for complex problem-solving. 

However, RL-based DeepSearch models frequently exhibit over-reliance on search, triggering tools unnecessarily due to difficulties in optimizing scalar rewards for both accuracy and efficiency.

\subsection{The Accuracy-Efficiency Trade-Off}

Recent advances in DeepSearch have yielded significant progress, yet these approaches grapple with a core dilemma: a see-saw trade-off between accuracy and efficiency. To address this challenge, many works focus on balancing reasoning accuracy with computational efficiency and resource utilization. For instance, DeepRAG~\cite{deeprag} employs a binary tree to construct retrieval routes, integrating supervised fine-tuning (SFT) and preference alignment to optimize both dimensions; Self-DC~\cite{self-dc} proposes a prompting framework that leverages LLMs' confidence scores to strategically decide tool invocation timing, reducing unnecessary searches to boost efficiency without sacrificing excessive accuracy; building on this, SMART~\cite{smart} introduces a refined fine-tuning approach using curated datasets to distinguish queries requiring external knowledge (needing search for accuracy) from those answerable via parametric memory (avoiding redundant search for efficiency). Similarly, SmartCal~\cite{smartcal} and Adaptive-RAG~\cite{adaptive-rag} extend such balance-seeking efforts to diverse tool spaces and reasoning scenarios.

Nevertheless, most of these methods fail to fully resolve the see-saw trade-off. Critically, they are not designed for the RL-driven autonomous tool invocation setting. Typically relying on extensive prompt engineering or manual annotations, they use static decision rules or predefined confidence thresholds, limiting their adaptability to dynamic queries and being inherently incompatible with DeepSearch’s RL training.

\subsection{Experience Memory Utilization}

In the evolution of existing large language models, experiential memory endows them with self-evolution capabilities, primarily manifested through experience-based learning mechanisms. 
For example, ORPO~\cite{orpo} enables models to iteratively rewrite prompts based on feedback from prior outputs. ADO~\cite{ado} introduces DSP, imposing semantic constraints on proposed prompts to identify optimal ones. ProTeGi~\cite{protegi} generates natural-language "corrections" applied as prompt edits, mimicking textual gradient descent. Reflexion~\cite{reflexion} establishes a self-reflection framework, allowing models to learn from failures and refine future behavior. PromptAgent~\cite{prompt-agent} frames prompt optimization as Monte Carlo Tree Search to strategically navigate instruction space and SPO~\cite{spo} creates a fully self-contained loop where the model generates its training data and uses pairwise preference comparison on its outputs to refine the prompt. 
Collectively, these works illustrate how LLMs can autonomously enhance their prompting policies, thereby boosting overall performance~\cite{cross-task-experience}. 

Our method LightSearcher leverages experiential memory to facilitate the evolution of DeepSearch. It extends self-evolution by integrating experiential memory from contrastive analysis of reasoning trajectories. This enables our model to distill interpretable textual guidance, optimizing tool invocation to balance efficiency and accuracy in deep reasoning systems.

\section{Preliminary}

\subsection{DeepSearch Framework}
DeepSearch extends traditional large language models' inference by dynamically searching external knowledge during the reasoning process~\cite{rag-reasoning-review}. Unlike static retrieval approaches that fetch information once before generation, DeepSearch enables iterative information search as reasoning progresses—embodying its core paradigm of synergizing real-time knowledge retrieval with stepwise inference.

Formally, given a query $q$, the objective is to generate an accurate answer $y$ by strategically invoking Search Tools as required. This is accomplished via an iterative paradigm comprising three core operations, i.e., \textbf{Reasoning, Decision}, and \textbf{Search}.

\textbf{Reasoning}: At each step $t$, the model generates reasoning content based on the current context:
\begin{equation}
r_t = \text{Reason}(c_t, q),
\end{equation}
where $c_t$ is the context containing previous reasoning steps and retrieved information.

\textbf{Decision}: The model decides whether to continue reasoning, search for external information, or generate the final answer:
\begin{equation}
d_t \in \{\text{search}, \text{continue}, \text{answer}\}.
\end{equation}

\textbf{Search}: When $d_t = \text{search}$, the model formulates a search query and retrieves relevant information:
\begin{equation}
s_t = \text{Search}(\text{Generate\_Query}(r_t, c_t, q), \mathcal{K}),
\end{equation}
where $\mathcal{K}$ represents the external knowledge base.

This iterative process continues until the model decides to generate the final answer. A complete reasoning trajectory is represented as:
\begin{equation}
\tau = (q, (r_1, d_1, s_1), (r_2, d_2, s_2), ..., (r_T, d_T, s_T), y),
\label{eq:trajectory1}
\end{equation}
where $r_i$ is the reasoning step, $d_i$ is the decision, $s_i$ is the result from Search Tool (empty if $d_i \neq \text{search}$), and $y$ is the final answer.

\subsection{Reinforcement Learning in DeepSearch}
Effective DeepSearch training relies on reinforcement learning (RL) to optimize decisions for invoking Search Tools strategically. 
The RL framework enables the model to acquire adaptive search behaviors through iterative policy refinement based on the reward function.

\subsubsection{Policy Rollout in RL}

During training, we sample multiple trajectories from the current policy $\pi_\theta$ to estimate the policy gradient. For each query $q$, the rollout process generates a complete trajectory:

\begin{equation}
\text{Rollout}(\pi_\theta, q) =(q, (r_1, d_1, s_1), ..., (r_T, d_T, s_T), y).
\end{equation}

At each step $t$, the model samples an action from the policy distribution:
\begin{equation}
d_t \sim \pi_\theta(d_t | s_t, q),
\end{equation}
where $s_t$ represents the current state containing the reasoning history. The rollout continues until the model generates a termination action ($d_T = \text{answer}$), producing the final answer $y$.

\subsubsection{Reward Function in RL}

Each sampled trajectory is evaluated using a multi-objective reward function:
\begin{equation}
R(\tau) = \mathcal{R}(F1(\tau), \text{Format}(\tau),\text{Tool}(\tau)).
\end{equation}

The policy is then updated using the collected trajectories to maximize expected reward:
\begin{equation}
\nabla_\theta J(\theta) = \mathbb{E}_{\tau \sim \pi_\theta} \left[\sum_{t=1}^{T} \nabla_\theta \log \pi_\theta(d_t | s_t, q) \cdot R(\tau)\right].
\end{equation}

This sampling-based approach explores different search strategies and learn from the outcomes, gradually improving its ability to balance accuracy and efficiency in deep reasoning models.

\section{Methodology}

\begin{figure*}[t]
    \centering
    \includegraphics[width=0.9\linewidth]{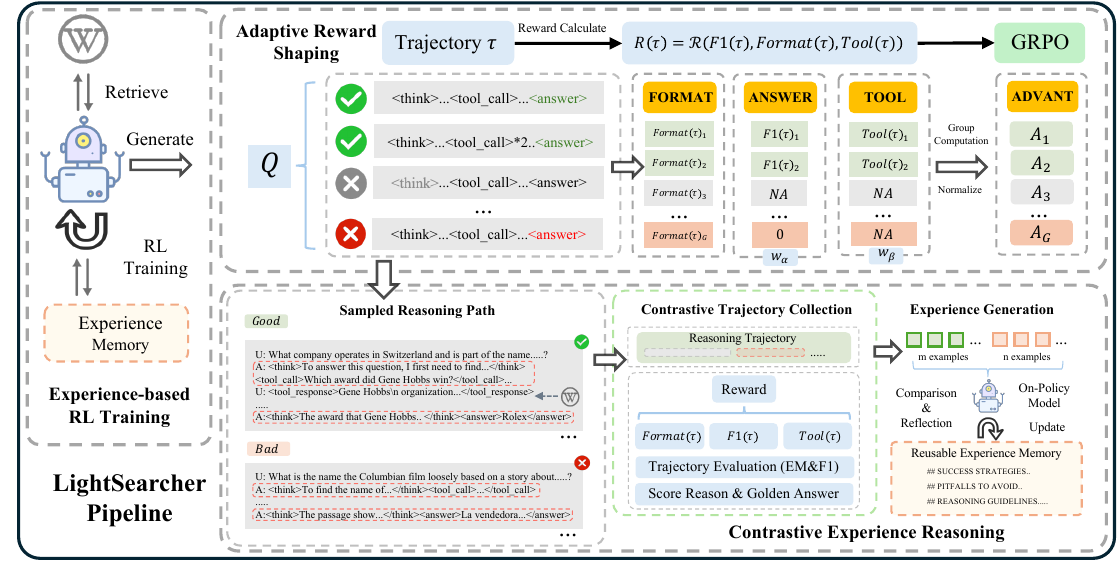}
    \caption{Overview of the LightSearcher framework pipeline. It contains three core modules: Contrastive Experience Reasoning, Adaptive Reward Shaping, and Experience-based RL training modules.}
    \label{fig:lightsearcher_overview}
\end{figure*}

We introduce the overview pipeline of our LightSearcher framework. LightSearcher addresses the accuracy-efficiency trade-off problem in deep reasoning models by incorporating experience learning and adaptive multi-objective rewards design.

\subsection{LightSearcher Pipeline}

LightSearcher is a search-enhanced reasoning model optimized by reinforcement learning. The overview architecture of LightSearcher is shown in Figure~\ref{fig:lightsearcher_overview}. It comprises three core modules:
\begin{itemize}
\item  \textbf{Contrastive Experiential Reasoning}: This module dynamically utilizes summaries of contrastive experience learned from past reasoning trajectories for RL optimization. 

\item \textbf{Adaptive Reward Shaping}: This module balances accuracy and efficiency by penalizing excessive search tool calls only in correct-answer scenarios, e.g., reducing invocations for queries resolvable with minimal tools.

\item \textbf{Experience-based RL Training}: This module integrates accumulated experiential memory and few-shot examples into prompts during the rollout process to enhance policy optimization.
 \end{itemize}

\subsection{Contrastive Experience Reasoning}

The core innovation of LightSearcher resides in its contrastive experiential reasoning mechanism. By leveraging the experiential memory of past reasoning trajectories based on accuracy, it automatically transforms good implicit reasoning experiences into explicit, interpretable textual guidance, thereby directing and optimizing the generation of future reasoning paths.

\subsubsection{Contrastive Trajectory Collection}

In training iteration $t$, we collect a set of reasoning trajectories $\mathcal{T}_t = \{\tau_1, \tau_2, ..., \tau_N\}$ generated by the current policy $\pi_\theta$. A trajectory $\tau$ is defined as: 
\begin{equation}
\tau = (q, (r_1, d_1, s_1), ..., (r_j, d_j, s_j), ..., (r_T, d_T, s_T), y),
\label{eq:trajectory2}
\end{equation}
where $r_j$ is the reasoning step, $d_j$ is the decision, and $s_j$ is the search result at step $j$ (see in Eq.~\ref{eq:trajectory1}).

We compute comprehensive reward scores for each trajectory using our multi-objective reward function (see in Eq.~\ref{eq:reward_function}):
\begin{equation}
R(\tau) = \mathcal{R}(F1(\tau), \text{Format}(\tau),\text{Tool}(\tau)).
\label{eq:reward_function}
\end{equation}
The trajectories are then categorized into contrastive groups based on their reward scores:
\begin{align}
\text{Good}(\tau) &= \{\tau \in \mathcal{T}_t : R(\tau) = 1\}, \\
\text{Bad}(\tau) &= \{\tau \in \mathcal{T}_t : R(\tau) < \theta_r\},
\label{eq:good_bad_trajectory}
\end{align}
where good trajectories achieve the most idealized outcome, i.e., the reward score =1, and bad trajectories have reward values that fall below a threshold $\theta_r = 0.3$.

\subsubsection{Experience Generation}

To generate meaningful experiences, we first augment each trajectory with an explicit explanation of its performance, formulated as:
\begin{equation}
\text{Sum}(\tau) = (\tau, F1(\tau), R(\tau), \text{Explanation}(\tau)),
\label{eq:summarize_trajectory}
\end{equation}
where $\text{Explanation}(\tau)$ provides explicit textual explanations for the reward assignment of a reasoning trajectory $\tau$ with consideration of its response quality, i.e., F1 score.
The summarized information $\text{Sum}(\tau)$ is further contrastively analyzed between high and low-quality trajectories. 
The experience generation process is then formalized as:
\begin{equation}
\text{Experience} = \text{LLM} (\text{Sum}({Good(\tau))}, \text{Sum}({Bad(\tau))}),
\label{eq:experience_generation}
\end{equation}
where ${Good(\tau)}$ and ${Bad(\tau)}$ represent collections of trajectories with high-reward and low-reward (defined in Eq.~\ref{eq:good_bad_trajectory}), respectively. 

The generated $\text{Experience}$ takes the form of natural language guidelines (Cases see in Fig.~\ref{fig:experience_case}) that explicitly describe effective reasoning patterns by learning from the comparisons of reasoning trajectories with distinct qualities. The experience is updated every 5 steps, maintaining a dynamic experiential memory bank.

\subsection{Adaptive Reward Shaping}
\label{sec:adaptive_reward_shaping}
Upon the integration of search tools, reinforcement learning-based reasoning models exhibit a propensity for excessive tool invocations, driven by the imperative to maximize accuracy~\cite{r1-searcher,research}.
To address this problem, existing studies incorporate a tool-use penalty term into the reward function~\cite{otc-po}, defined as:
\begin{equation}
  R(\tau) = \mathcal{R}(\text{F1}(\tau), \text{Format}(\tau), \text{Tool}(\tau)),
  \label{eq:reward}
\end{equation}
where $\text{F1}$, $\text{Format}$, $\text{Tool}$ denote the rewards for answer accuracy, format compliance, and tool usage, respectively. $\mathcal{R}$ is the aggregation function that combines multiple objectives for optimization.

Existing RL methods employ simple weighted summation for reward adjustments and mostly prioritize response accuracy as the primary objective, causing models to increase retrieval frequency to ensure accuracy, thereby leading to redundant retrieval and efficiency degradation. 
To address this issue, we propose a novel adaptive reward shaping mechanism that penalizes excessive tool calls only for queries that can be correctly answered using the model's intrinsic parametric knowledge alone.

Given a query $q$, we record the minimal tool usage number $n$ for a correct answer over past training trajectories. $\theta_t$ is the threshold of the F1 score for determining whether the answer is correct. When a trajectory achieves an F1 exceeding $\theta_t$, we record the minimum number of tool invocations $n$ as the baseline for tool calls. For subsequent trajectories in the same problem, we employ the following function to penalize excessive search tool usage:
\begin{equation}
  \text{Tool}(\tau) = \begin{cases} e^{-\lambda \cdot \max(0, m - n)}, & \text{if } \text{F1}(\tau) \geq \theta_t, \\
  0, & \text{if } \text{F1}(\tau) < \theta_t,
  \end{cases}
  \label{eq:tool_reward}
\end{equation}
where $m$ is the actual number of tool calls in the current trajectory, $n$ is the recorded minimum tool calls, and $\lambda$ is a hyper-parameter controlling the penalty strength. This function applies a smooth, exponential decay based on tool usage that exceeds the established baseline $n$ to ensure the efficiency of tool calls.

As for the reward of format $\text{Format}(\tau)$, we set it as a fundamental compliance constraint and use a binary indicator to calculate, defined as: 
\begin{equation}
  \text{Format}(\tau) = \begin{cases}
    -1, & \text{Incorrect format}, \\
    0, & \text{Correct format},
  \end{cases}
\end{equation}
where a correct format requires responses to begin with <think> tags for reasoning, end with </think>, and follow with either <answer> or <tool\_call> tags.

The accuracy component $\text{F1}(\tau)$ directly represents the task-level F1 score, serving as the primary accuracy measure. The overall reward function in our lightSearcher is defined as:
\begin{equation}
 R(\tau) = \begin{cases}
    -1, & \text{if } \text{Format}(\tau) = -1, \\
    W_{\alpha} \text{F1}(\tau) + W_{\beta} \cdot \text{Tool}(\tau), & \text{if } \text{Format}(\tau) = 0,
    \end{cases}
    \label{eq:overall_reward}
\end{equation}
where $W_\alpha$ and $W_\beta$ are hyperparameters that adjust the importance of rewards for accuracy and tool calls. This adaptive adjustment strategy enables our model to gradually learn efficient tool invocation behaviors while ensuring performance.

\subsection{Experience-based RL Training}

After integrating the experiential memory into the reinforcement learning process, we adopt GRPO (Group Relative Policy Optimization)~\cite{grpo} to train our model.
In each training iteration, all experience from the experiential memory bank 
is integrated into the model's input prompt to provide comprehensive guidance. Additionally, we randomly select a high-quality trajectory from previous successful cases to serve as few-shot examples, defined as:
\begin{equation}
\text{Few-shot}(\tau) = \text{RandomSample}(\{\tau : \mathcal{R}(\tau) \geq \theta_{r}\}).
\end{equation}

The augmented prompt template combines all experiential memory with the few-shot example, defined as:
\begin{equation}
\text{Prompt}_{\text{aug}} = \{\text{Instructions}, \text{Experience}, \text{Few-shot}(\tau), q\},
\end{equation}
where $\text{Instructions}$ provides the basic task description and format requirements,
$\text{Experience}$ is the contains the accumulated experiential memory (Eq.~\ref{eq:experience_generation}), $\text{Few-shot}(\tau)$ is a sampled high-quality trajectory example, and $q$ is the query to be answered.

The GRPO training process is formulated as:
\begin{equation}
\mathcal{L}_{\text{GRPO}} = \mathbb{E}_{\tau \sim \pi_\theta(\cdot|\text{Prompt}_{\text{aug}})} \left[ \sum_{t=0}^{T} \log \pi_\theta(a_t|s_t, \text{Prompt}_{\text{aug}}) \cdot A_t \right],
\end{equation}
where $A_t$ represents the advantage function computed using our multi-objective reward function $R(\tau)$ (Eq.~\ref{eq:overall_reward}). This experience-guided training enables the model to learn more efficiently by leveraging historical insights while maintaining exploration capability through the stochastic policy optimization process.

\section{Experiments}

\subsection{Experimental Settings}

\subsubsection{Datasets}
We conduct experiments on four representative datasets in search-enhanced reasoning research.
\begin{itemize}
    \item \textbf{Natural Questions (NQ)}~\cite{nq}: a large-scale QA dataset with real user questions and Wikipedia passages, treated as out-of-domain test set with 1,000 randomly sampled instances.
    \item \textbf{HotpotQA}~\cite{hotpotqa}: a dataset requiring reasoning over multiple supporting documents, treated as out-of-domain test set with 1,000 randomly sampled instances.
    \item \textbf{Musique}~\cite{musique}: a multi-hop QA dataset via single-hop question composition, used as in-domain test set with 1,000 randomly sampled instances.
    \item \textbf{2WikiMultihopQA}~\cite{2wiki}: a large-scale multi-hop reading comprehension dataset, used as in-domain test set with 1,000 randomly sampled instances.
\end{itemize}

\begin{figure*}[t]
\centering
\includegraphics[width=0.9\textwidth]{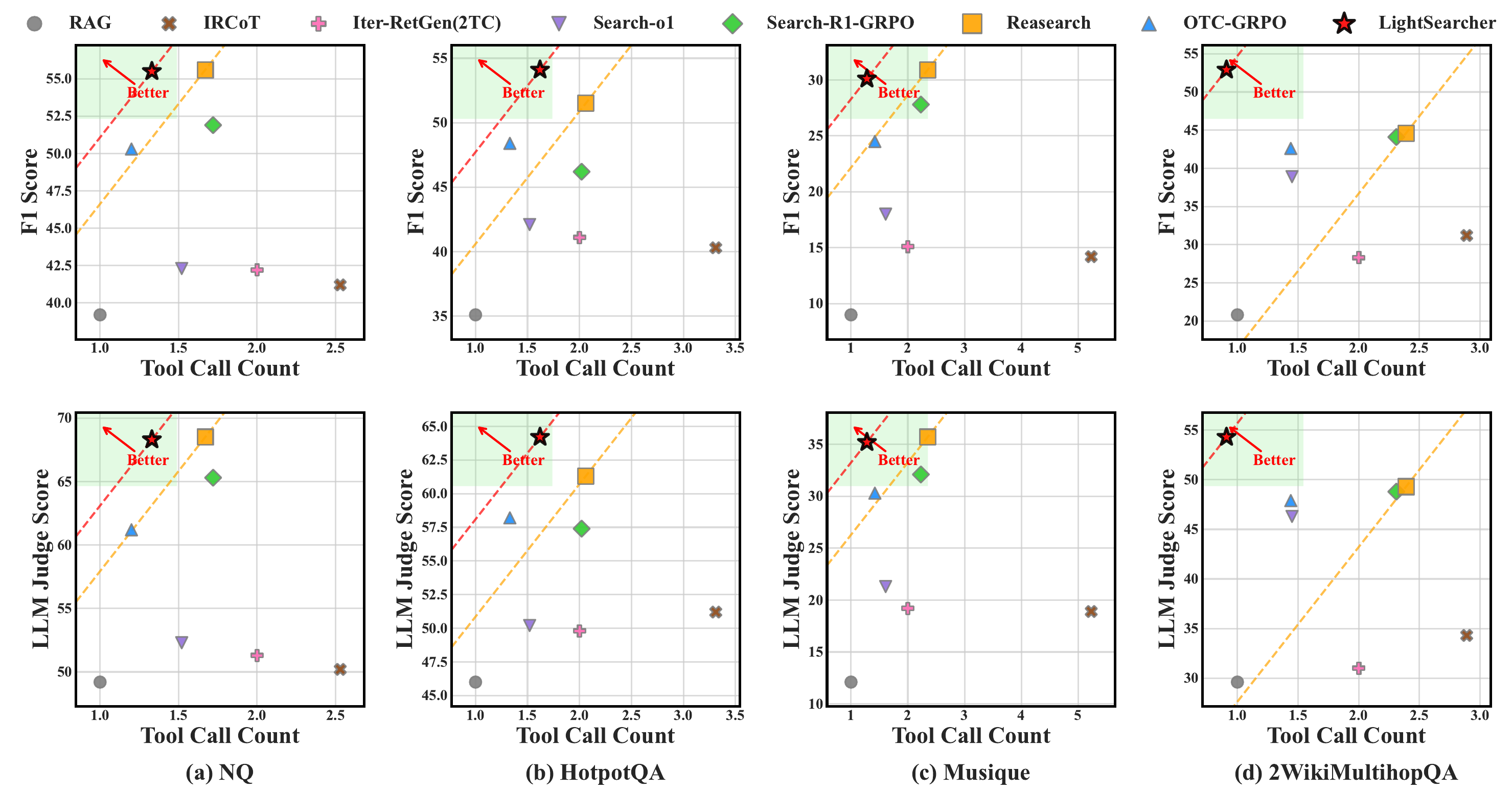}
\caption{Performance comparison of LightSearcher and baseline methods across four datasets on Qwen2.5-7B-Instruct.
The figure presents the relationship between F1 score, LLM judge, and search tool calls, demonstrating that our model achieves the optimal balance in terms of effectiveness and search overhead.
 }
\label{fig:main_results}
\end{figure*}

\subsubsection{Evaluation Metrics}

Evaluation metrics include: (1) F1 score, measuring the overlap between answers and ground truth; (2) Exact Match (EM), requiring perfect answer matching; (3) LLM-as-a-Judge (LMJ), using DeepSeek-V3 to automatically evaluate answer quality; (4) Tool Call count (TC), measuring models' efficiency.

\subsubsection{Baseline Methods}
We compare LightSearcher against a comprehensive set of baselines spanning different methodological approaches to search-enhanced reasoning. For \textbf{Iterative methods}, we make comparisons with \emph{Naive RAG}~\cite{rag}, a standard retrieval-augmented generation baseline that performs single-step retrieval followed by answer generation. \emph{IRCoT}~\cite{ircot} employs chain-of-thought reasoning to iteratively construct intermediate logic chains for multi-hop retrieval. \emph{Iter-RetGen}~\cite{iter-retgen} reconstructs subsequent queries using intermediate answers to resolve multi-hop dependencies with a fixed two-step retrieval constraint. \emph{Search-o1}~\cite{search-o1} is a prompting-based method that enables reasoning models to perform strategic tool invocation during inference through step-by-step reasoning. For \textbf{Reinforcement Learning (RL)} methods, we include \emph{Search-R1}~\cite{search-r1}, which utilizes GRPO training to optimize tool invocation strategies, and Research~\cite{research}, which employs reinforcement learning with outcome-based rewards to improve multi-hop reasoning performance. \emph{OTC-GRPO}~\cite{otc-po} represents a standard GRPO implementation with outcome-based rewards specifically designed for tool-augmented reasoning tasks, serving as a direct comparison to our experience-enhanced approach.
Different from them, LightSearcher presents a reinforcement learning framework that optimizes search-enhanced reasoning through textual experiential memory. We employ experience-enhanced learning to guide tool invocation strategies rather than solely relying on outcome-based rewards.

\subsubsection{Implementation Details.}
We conducted experiments on two base models: Qwen2.5-3B-Instruct and Qwen2.5-7B-Instruct. 
We use the codebase of Reasearch~\cite{research}, which employed Verl~\cite{verl} as the foundational framework for reinforcement learning and adopted the GRPO training method. For the iterative retrieval methods, we utilized the FlashRAG framework~\cite{flashrag}, a standardized and widely adopted toolkit. The retriever was implemented using E5-base-v2~\cite{e5}, with the KLIT 2018 Wikipedia corpus~\cite{kilt} serving as the knowledge base for retrieval. 
Following the data mixture strategy in existing literature~\cite{search-r1, r1-searcher}, we sampled 3,000 examples from the Musique dataset and 4,000 examples from the Wikipedia dataset for training. 
During training, we train with a batch size of 192, rollout number set to 12, a learning rate of 1e-6, and trained for 10 epochs. In our reward function, the F1 score threshold $\theta_r$ in Eq.~\ref{eq:tool_reward} is set to 0.8.
The weighting parameters $W_{\alpha}$ and $W_{\beta}$ in Eq.~\ref{eq:overall_reward} are both set to 0.5, balancing the contributions of different objectives (such as accuracy and efficiency) in the overall reward. The parameter $\lambda$ in Eq.\ref{eq:tool_reward} is set to 0.75 for moderate decay.

\subsection{Main Results}

Performance comparisons of LightSearcher and baseline methods across four datasets are shown in Fig.~\ref{fig:main_results}. 
The Figure illustrates F1 and LLM Judge scores along with search tool call counts on the Qwen2.5-7B-Instruct model. We have the following observations:

(1) \textbf{Overall Performance Superiority}:
Overall, RL-based reasoning models (i.e., Search-R1, OTC, ReSearch, LightSearcher) achieve higher accuracy than Iterative methods (i.e., RAG, IRCoT, Iter-RetGen),  demonstrating that automatically learning search-enhanced reasoning trajectories leads to better performance. Among RL-based methods, our LightSearcher and ReSearch demonstrate comparable top-tier performance. LightSearcher achieves substantial improvements over the SOTA baseline (ReSearch) on HotpotQA and 2WikiMultihopQA, while showing slight performance decreases on NQ and Musque datasets. These results demonstrate that the experience-enhanced learning mechanism effectively guides the model toward more optimal reasoning strategies. 

(2) \textbf{Remarkable Efficiency Improvements}: 
The most significant advantage of LightSearcher lies in its efficiency improvements while maintaining competitive accuracy. While Naive RAG and OCT require fewer search tool calls, this is gained at the cost of model performance loss. Compared to the best-performing baseline Research, our LightSearcher achieves an
39.6\% average reduction of search tool calls across four datasets, showing a significant computational efficiency gain.
This efficiency gain stems from the contrastive experience generation mechanism in LightSearcher, which identifies patterns in successful trajectories with fewer tools, explicitly teaching strategic restraint in tool invocation.

(3) \textbf{Cross-Domain Generalization Capability}: LightSearcher exhibits strong generalization performance across in-domain and out-of-domain datasets. LightSearcher is trained on Musique and 2WikiMultihopQA datasets. On out-of-domain datasets (NQ and HotpotQA), LightSearcher maintains competitive performance while significantly reducing tool usage. This generalization capability suggests that the learned experiences capture fundamental reasoning strategies rather than task-specific patterns.

\section{Experimental Analysis}
In this section, we first analyze the role of each module of LightSearcher in the ablation study, conducting an in-depth analysis on model efficiency, and providing case studies for empirical evolution.

\begin{table}
  \caption{
  Ablation study evaluating component contributions in LightSearcher on HotpotQA dataset using Qwen2.5-7B-Instruct model.
  }
  \label{tab:ablation_results}
  \centering
   \small
  \begin{tabular}{lccc}
    \toprule
    \textbf{Model Variant} & \textbf{F1} & \textbf{LMJ} & \textbf{TC} \\
    \midrule
    LightSearcher & 54.1  & 64.2 & 1.62 \\
    \midrule
    w/o Exp (c1) & 50.2 ($\textbf{7.2\%} \downarrow$) & 60.4 ($\textbf{5.9\%}\downarrow$) & 1.18\\
    w/o Few-shot (c2) & 51.5 ($4.8\% \downarrow$) & 61.3 ($4.5\% \downarrow$) & 1.33\\
    w/o Adaptive Rewards (c3) & 54.5 ($0.7\% \uparrow$) & 65.3 ($1.7\% \uparrow$) & 2.06 \\ \hline
    w/o Exp \& Few-shot (-c1,c2) & 49.3 ($8.8\% \downarrow$) & 58.1 ($9.5\% \downarrow$) & 1.03\\
    Only GRPO (-c1,c2,c3) & 51.3 ($5.1\% \downarrow$) & 63.1 ($1.7\% \downarrow$) & 2.16\\
    \bottomrule
  \end{tabular}
\end{table}

\subsection{Ablation Study}

To address the effectiveness of each component in LightSearcher, we conduct ablation studies by systematically removing key components from our framework. We evaluate each variant on the HotpotQA dataset using Qwen2.5-7B-Instruct as the base model, analyzing both performance metrics and efficiency indicators. 
As shown in Table~\ref{tab:ablation_results}, we can see that:

(1) Experiential Memory is the most critical component. Removing the experiential memory component (w/o Exp) results in the most significant performance degradation, with F1 score dropping by 7.2\%  and LMJ score decreasing by 5.9\%.
Notably, the tool call count decreases to 1.18, showing that without experience guidance, the model turns overly conservative and neglects to retrieve essential information as needed. This underscores the vital role of textual experience in optimizing accuracy and making strategic tool invocation decisions.

(2) Eliminating the few-shot inspiration mechanism (w/o Few-shot) leads to decreases on F1 and LMJ scores.
This indicates that while specific reasoning path examples provide valuable guidance for query construction and reasoning structure, the impact is less pronounced than the experience memory mechanism. The model can partially compensate through accumulated textual experience, but it loses the fine-grained procedural guidance that few-shot examples provide.

(3) Removing adaptive rewards (w/o Adaptive Rewards) results in a surprisingly marginal improvement in model accuracy metrics, yet induces a substantial escalation in search tool invocations to 2.06 (a 27\% rise relative to LightSearcher). 
This component appears to be primarily responsible for efficiency optimization rather than accuracy improvement. The static reward weighting fails to adaptively balance accuracy-efficiency trade-offs during training, leading to suboptimal tool invocation strategies.

\begin{figure}
  \centering
  \includegraphics[width=0.5\textwidth]{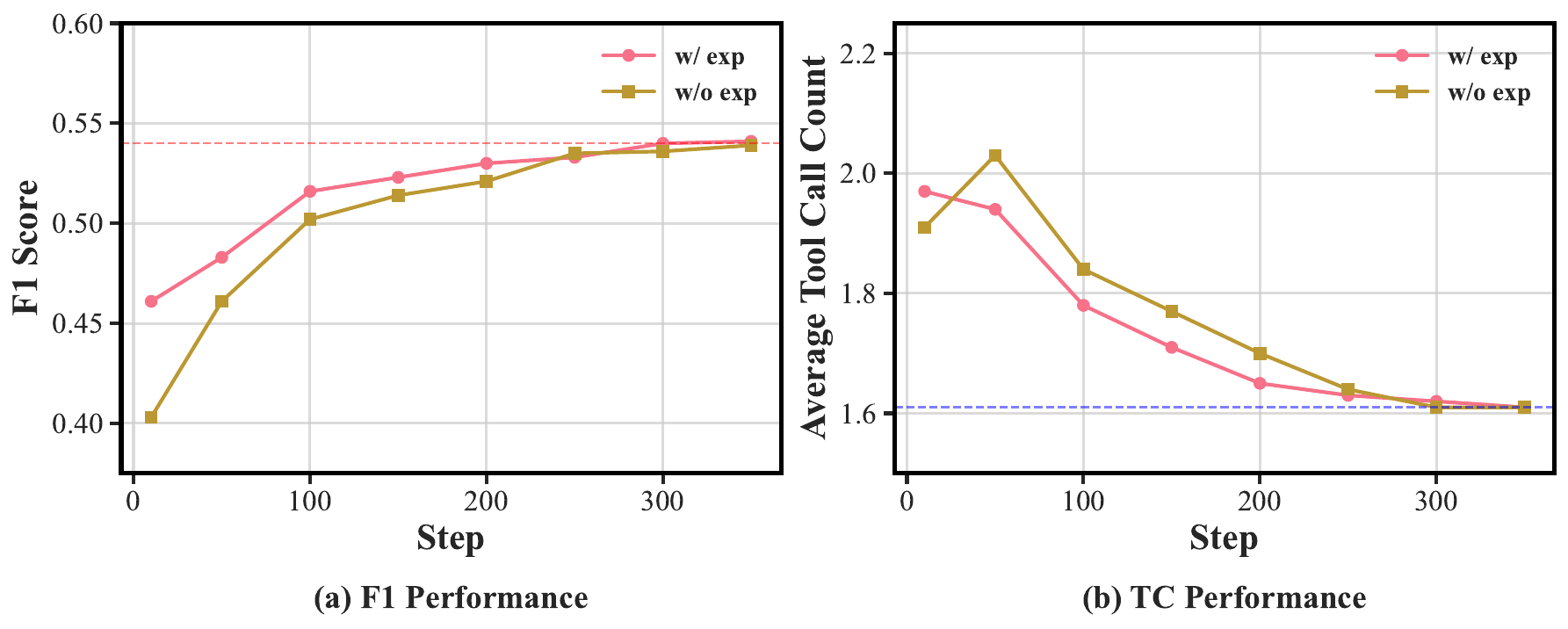}
  \caption{
  The impact of experience incorporation in inference on F1 score and Tool calls across training steps.
 }
  \label{fig:ablation_results_inference}
\end{figure}

\begin{figure}
  \centering
  \includegraphics[width=\linewidth]{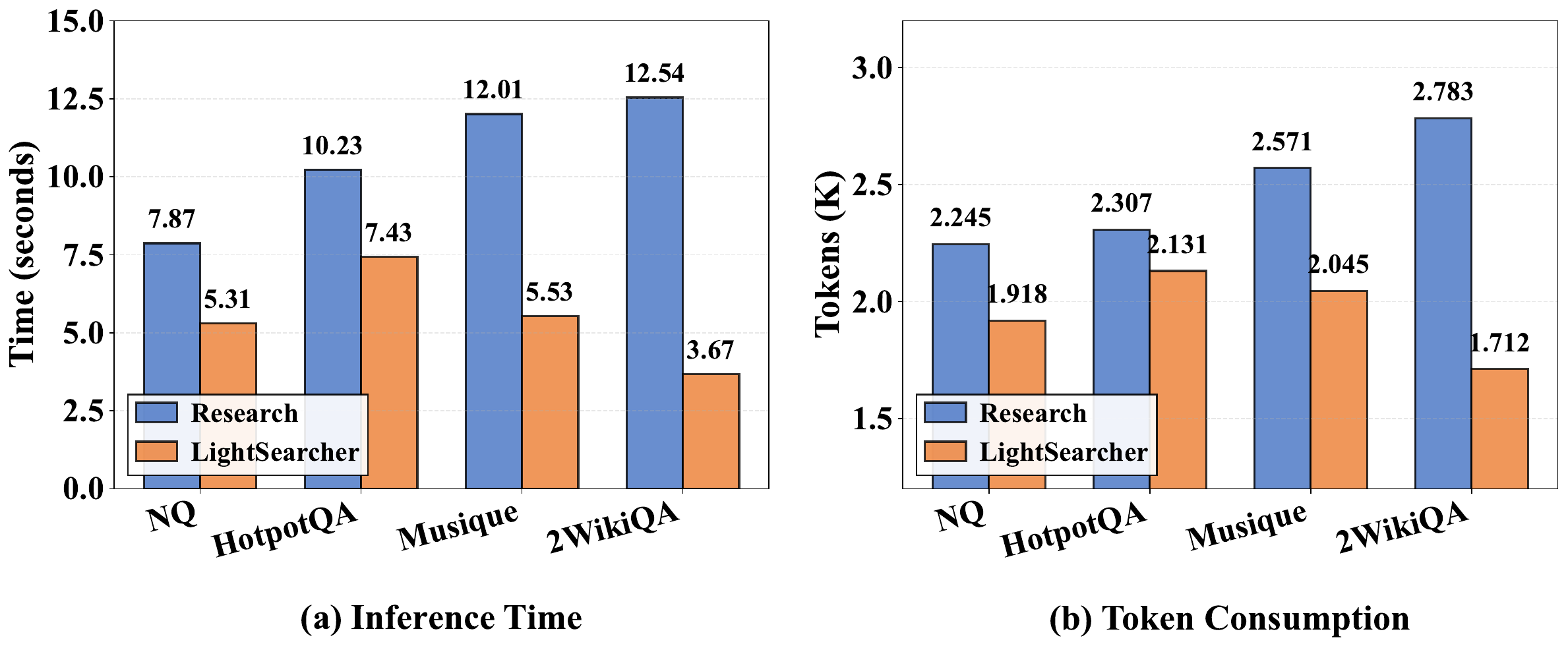}
  \caption{ Computational efficiency comparison between LightSearcher and ReSearch, evaluating Inference Time and Token Consumption.
}
  \label{fig:computational_efficiency}
\end{figure}

\begin{figure*}
  \centering
  \includegraphics[width=0.9\textwidth]{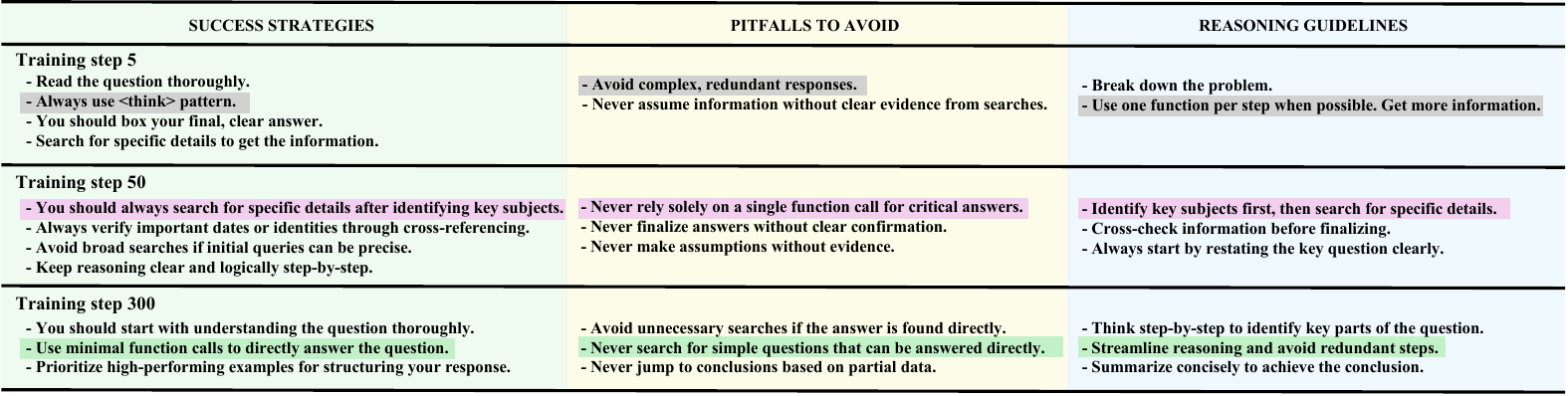}
  \caption{
  Evolution of generated experience memory throughout training, illustrating progression from early to advanced stages on three prompt categories of experience: Success Strategies, Pitfalls to Avoid, and Reasoning Guidelines.
  }
  \label{fig:experience_case}
\end{figure*}

\subsection{Inference Efficiency Analysis}
This subsection analyzes the impact of incorporating experiential memory on model inference performance and analyzes the inference efficiency in terms of inference time and token consumption.

\subsubsection{Inference with Experiential Memory}
During the RL training process, integrating reasoning experiences significantly enhances the model's performance. However, since incorporating these experiences during inference introduces additional computational overhead, we further examine the necessity of their inclusion in the inference phase.
As shown in Fig.~\ref{fig:ablation_results_inference}, we evaluate the impact of experience texts on F1 and tool call counts during inference across various training steps. The results reveal that reasoning experiences exert a more pronounced influence on inference performance in the early training stages. 
As training progresses, the impact of experience texts diminishes, as the model converges to a stable policy. Consequently, our model eliminates the need for these experiences during inference, thereby avoiding unnecessary computational costs.

\subsubsection{Inference Time and Token Consumption}
To assess the computational cost reductions attained by LightSearcher,
we conduct a comparison with the SOTA baseline ReSearch, evaluating their inference times and token consumption across various datasets.
As illustrated in Fig.~\ref{fig:computational_efficiency}, LightSearcher achieves average reductions of 48.6\% in inference time (Time) and 21.2\% in token consumption (Tokens) relative to ReSearch across four datasets.
These improvements in inference efficiency derive from our model's strategic tool invocation decisions and the minimization of redundant retrieval operations, guided by accumulated experiential memory, thereby substantiating the efficiency advantages of our approach.

\subsection{Adaptive Retrieval Boosts Efficiency}

Our experience-based adaptive reward mechanism penalizes search tool invocations for correctly answered questions, thereby reducing redundant searches in those cases.
To further investigate differences in tool invocation patterns between successful and failed reasoning attempts, we analyzed the distribution of tool call counts for correct versus incorrect answers across various datasets.
Table~\ref{tab:tool_invocation_analysis} presents the average tool call count per question for both correct and incorrect cases in our evaluation datasets.
Our method exhibits significantly lower average tool call counts for both EM=0 and EM=1 compared to ReSearch.
LightSearcher demonstrates robust self-adaptation to diverse queries: correct answers (EM=1) require an average of 9.71\% fewer tool calls (i.e., $\Delta\%$ =9.71) than incorrect ones (EM=0), compared to only a 6.27\% difference for ReSearch. This suggests that LightSearcher adopts a more flexible retrieval strategy.
This disparity is particularly pronounced in datasets like 2WikiMultihopQA (2WikiQA) (16.65\%) and NQ (15.65\%), which exhibit higher overall question accuracy than the Musique dataset. In contrast, on the Musique dataset, LightSearcher's retrieval attempts for incorrectly answered questions (EM=0) are fewer than for correct ones, a reversal of the trend seen in other datasets.
This may arise from the dataset's exceptional difficulty, evidenced by its lowest accuracy (Fig.~\ref{fig:main_results}), leading the model to curtail excessive retrieval efforts on incorrectly answered questions.
Our model automatically minimizes such retrievals while sustaining high overall accuracy, highlighting its capacity to improve precision alongside optimized efficiency.

\begin{table}
  \caption{
  Analysis of search Tool Calls in correct versus incorrect answer cases across multiple datasets, highlighting the distinct patterns in the tool usage efficiency of our LightSearcher. $\Delta\%$ represents the average percentage reduction in Tool Calls required for correct answers (EM=1) compared to incorrect ones (EM=0).
  }
  \label{tab:tool_invocation_analysis}
  \centering
  \small
  \setlength{\tabcolsep}{1.8pt}
  \begin{tabular}{lcccccccc}
    \toprule
    \multirow{2}{*}{\textbf{Dataset}}
    & \multicolumn{3}{c}{\textbf{LightSearcher}}
    & & \multicolumn{3}{c}{\textbf{Research}} \\
    \cmidrule(lr){2-4} \cmidrule(lr){6-8}
    & \textbf{TC(EM=1)} & \textbf{TC(EM=0)} & $\Delta$\%~$\uparrow$
    & & \textbf{TC(EM=1)} & \textbf{TC(EM=0)} & $\Delta$\%~$\uparrow$ \\
    \midrule
    NQ & 1.207 & 1.431 & 15.65 & & 1.583 & 1.729 & 11.05 \\
    HotpotQA & 1.535 & 1.76 & 12.78 & & 2.025 & 2.087 & 2.97 \\
    Musique & 1.343 & 1.259 & -6.67 & & 2.233 & 2.385 & 6.37 \\
    2WikiQA & 0.822 & 0.985 & 16.55 & & 2.305 & 2.442 & 5.64 \\
    \midrule
    \textbf{Average} & \textbf{1.227} & \textbf{1.359} & \textbf{9.71} & & 2.037 & 2.161 &  6.27 \\
    \bottomrule
  \end{tabular}
\end{table}

\subsection{Evolution of Experience Memory}
To better elucidate the role of explicit experiential memory, Fig.~\ref{fig:experience_case} presents excerpts from memories at training steps 5 (initial stage), 50 (early stage), and 300 (advanced stage) across three template categories: \emph{Success Strategies}, \emph{Pitfalls to Avoid}, and \emph{Reasoning Guidelines}.
Taking the experiences in \emph{Success Strategies} as an example: In the early training stages, the generated experiences primarily emphasize basic operational guidance, such as "Always use <think> pattern (step 5)," to ensure reasoning format correctness. As training advances, these experiences evolve to incorporate more sophisticated strategic patterns, for instance, "You should always search for specific details after identifying key subjects (step 50)." Ultimately, they pivot toward optimizing model efficiency, such as "Use minimal function calls to directly answer the question (step 300)."
Notably, this evolution mirrors the model's progressive capability enhancement: from an initial focus on format correctness, to validation of retrieved content, and finally to retrieval efficiency optimization. This incremental refinement of reasoning capabilities enables our model to effectively balance efficiency and performance.

\section{Conclusion}
LightSearcher advances search-enhanced reasoning by integrating contrastive experiential memory and adaptive reward shaping within an RL framework, enabling LLMs to strategically invoke tools while balancing accuracy and efficiency. By enabling models to learn experience from past reasoning trajectories, our method promotes more efficient and adaptive search tool-augmented AI systems, improving resource utilization across diverse tasks.
Despite these advancements, LightSearcher is currently suited to controlled reasoning scenarios and demands additional computational resources during RL training, which may present challenges for broader scalability. Additionally, while validated on multi-hop QA tasks, future directions include extending its application to domains like code synthesis and strategic planning.

\bibliographystyle{ACM-Reference-Format}
\bibliography{reference}

@article{measuring,
  title={Measuring massive multitask language understanding},
  author={Hendrycks, Dan and Burns, Collin and Basart, Steven and Zou, Andy and Mazeika, Mantas and Song, Dawn and Steinhardt, Jacob},
  journal={arXiv preprint arXiv:2009.03300},
  year={2020}
}

@article{medical-research,
  title={A study of generative large language model for medical research and healthcare},
  author={Peng, Cheng and Yang, Xi and Chen, Aokun and Smith, Kaleb E and PourNejatian, Nima and Costa, Anthony B and Martin, Cheryl and Flores, Mona G and Zhang, Ying and Magoc, Tanja and others},
  journal={NPJ digital medicine},
  volume={6},
  number={1},
  pages={210},
  year={2023},
  publisher={Nature Publishing Group UK London}
}

@inproceedings{llm-in-finance,
  title={Large language models in finance: A survey},
  author={Li, Yinheng and Wang, Shaofei and Ding, Han and Chen, Hang},
  booktitle={Proceedings of the fourth ACM international conference on AI in finance},
  pages={374--382},
  year={2023}
}

@article{rag,
  title={Retrieval-augmented generation for knowledge-intensive nlp tasks},
  author={Lewis, Patrick and Perez, Ethan and Piktus, Aleksandra and Petroni, Fabio and Karpukhin, Vladimir and Goyal, Naman and K{\"u}ttler, Heinrich and Lewis, Mike and Yih, Wen-tau and Rockt{\"a}schel, Tim and others},
  journal={Advances in neural information processing systems},
  volume={33},
  pages={9459--9474},
  year={2020}
}

@article{rag-survey,
  title={Retrieval-augmented generation for large language models: A survey},
  author={Gao, Yunfan and Xiong, Yun and Gao, Xinyu and Jia, Kangxiang and Pan, Jinliu and Bi, Yuxi and Dai, Yi and Sun, Jiawei and Wang, Haofen and Wang, Haofen},
  journal={arXiv preprint arXiv:2312.10997},
  volume={2},
  year={2023}
}

@article{rag-reasoning-review,
  title={Synergizing rag and reasoning: A systematic review},
  author={Gao, Yunfan and Xiong, Yun and Zhong, Yijie and Bi, Yuxi and Xue, Ming and Wang, Haofen},
  journal={arXiv preprint arXiv:2504.15909},
  year={2025}
}

@article{kg-retriever,
  title={Kg-retriever: Efficient knowledge indexing for retrieval-augmented large language models},
  author={Chen, Weijie and Bai, Ting and Su, Jinbo and Luan, Jian and Liu, Wei and Shi, Chuan},
  journal={arXiv preprint arXiv:2412.05547},
  year={2024}
}

@article{cross-task-experience,
  title={Cross-Task Experiential Learning on LLM-based Multi-Agent Collaboration},
  author={Li, Yilong and Qian, Chen and Xia, Yu and Shi, Ruijie and Dang, Yufan and Xie, Zihao and You, Ziming and Chen, Weize and Yang, Cheng and Liu, Weichuan and others},
  journal={arXiv preprint arXiv:2505.23187},
  year={2025}
}

@inproceedings{emotional-rag,
  title={Emotional RAG: Enhancing role-playing agents through emotional retrieval},
  author={Huang, Le and Lan, Hengzhi and Sun, Zijun and Shi, Chuan and Bai, Ting},
  booktitle={2024 IEEE International Conference on Knowledge Graph (ICKG)},
  pages={120--127},
  year={2024},
  organization={IEEE}
}

@article{memoryos,
  title={Memory OS of AI Agent},
  author={Kang, Jiazheng and Ji, Mingming and Zhao, Zhe and Bai, Ting},
  journal={arXiv preprint arXiv:2506.06326},
  year={2025}
}

@article{deepseek-r1,
  title={Deepseek-r1: Incentivizing reasoning capability in llms via reinforcement learning},
  author={Guo, Daya and Yang, Dejian and Zhang, Haowei and Song, Junxiao and Zhang, Ruoyu and Xu, Runxin and Zhu, Qihao and Ma, Shirong and Wang, Peiyi and Bi, Xiao and others},
  journal={arXiv preprint arXiv:2501.12948},
  year={2025}
}

@article{iter-retgen,
  title={Enhancing retrieval-augmented large language models with iterative retrieval-generation synergy},
  author={Shao, Zhihong and Gong, Yeyun and Shen, Yelong and Huang, Minlie and Duan, Nan and Chen, Weizhu},
  journal={arXiv preprint arXiv:2305.15294},
  year={2023}
}

@inproceedings{ircot,
  title={Interleaving Retrieval with Chain-of-Thought Reasoning for Knowledge-Intensive Multi-Step Questions},
  author={Trivedi, Harsh and Balasubramanian, Niranjan and Khot, Tushar and Sabharwal, Ashish},
  booktitle={Proceedings of the 61st Annual Meeting of the Association for Computational Linguistics (Volume 1: Long Papers)},
  pages={10014--10037},
  year={2023}
}

@article{search-r1,
  title={Search-r1: Training llms to reason and leverage search engines with reinforcement learning},
  author={Jin, Bowen and Zeng, Hansi and Yue, Zhenrui and Yoon, Jinsung and Arik, Sercan and Wang, Dong and Zamani, Hamed and Han, Jiawei},
  journal={arXiv preprint arXiv:2503.09516},
  year={2025}
}

@article{research,
  title={Research: Learning to reason with search for llms via reinforcement learning},
  author={Chen, Mingyang and Li, Tianpeng and Sun, Haoze and Zhou, Yijie and Zhu, Chenzheng and Wang, Haofen and Pan, Jeff Z and Zhang, Wen and Chen, Huajun and Yang, Fan and others},
  journal={arXiv preprint arXiv:2503.19470},
  year={2025}
}

@misc{otc-po,
      title={Acting Less is Reasoning More! Teaching Model to Act Efficiently}, 
      author={Hongru Wang and Cheng Qian and Wanjun Zhong and Xiusi Chen and Jiahao Qiu and Shijue Huang and Bowen Jin and Mengdi Wang and Kam-Fai Wong and Heng Ji},
      year={2025},
      eprint={2504.14870},
      archivePrefix={arXiv},
      primaryClass={cs.AI},
      url={https://arxiv.org/abs/2504.14870}, 
}

@article{r1-searcher,
  title={R1-searcher: Incentivizing the search capability in llms via reinforcement learning},
  author={Song, Huatong and Jiang, Jinhao and Min, Yingqian and Chen, Jie and Chen, Zhipeng and Zhao, Wayne Xin and Fang, Lei and Wen, Ji-Rong},
  journal={arXiv preprint arXiv:2503.05592},
  year={2025}
}

@article{self-rag,
  title={Self-rag: Learning to retrieve, generate, and critique through self-reflection},
  author={Asai, Akari and Wu, Zeqiu and Wang, Yizhong and Sil, Avirup and Hajishirzi, Hannaneh},
  year={2024},
  publisher={ICLR}
}

@article{search-o1,
  title={Search-o1: Agentic search-enhanced large reasoning models},
  author={Li, Xiaoxi and Dong, Guanting and Jin, Jiajie and Zhang, Yuyao and Zhou, Yujia and Zhu, Yutao and Zhang, Peitian and Dou, Zhicheng},
  journal={arXiv preprint arXiv:2501.05366},
  year={2025}
}

@article{nq,
  title={Natural questions: a benchmark for question answering research},
  author={Kwiatkowski, Tom and Palomaki, Jennimaria and Redfield, Olivia and Collins, Michael and Parikh, Ankur and Alberti, Chris and Epstein, Danielle and Polosukhin, Illia and Devlin, Jacob and Lee, Kenton and others},
  journal={Transactions of the Association for Computational Linguistics},
  volume={7},
  pages={453--466},
  year={2019},
  publisher={MIT Press One Rogers Street, Cambridge, MA 02142-1209, USA journals-info~…}
}

@inproceedings{hotpotqa,
  title={HotpotQA: A Dataset for Diverse, Explainable Multi-hop Question Answering},
  author={Yang, Zhilin and Qi, Peng and Zhang, Saizheng and Bengio, Yoshua and Cohen, William and Salakhutdinov, Ruslan and Manning, Christopher D},
  booktitle={Proceedings of the 2018 Conference on Empirical Methods in Natural Language Processing},
  pages={2369--2380},
  year={2018}
}

@article{musique,
  title={MuSiQue: Multihop Questions via Single-hop Question Composition},
  author={Trivedi, Harsh and Balasubramanian, Niranjan and Khot, Tushar and Sabharwal, Ashish},
  journal={Transactions of the Association for Computational Linguistics},
  volume={10},
  pages={539--554},
  year={2022}
}

@inproceedings{2wiki,
  author       = {Xanh Ho and
                  Anh{-}Khoa Duong Nguyen and
                  Saku Sugawara and
                  Akiko Aizawa},
  title        = {Constructing {A} Multi-hop {QA} Dataset for Comprehensive Evaluation
                  of Reasoning Steps},
  booktitle    = {{COLING}},
  pages        = {6609--6625},
  publisher    = {International Committee on Computational Linguistics},
  year         = {2020}
}

@article{ppo,
  title={Proximal policy optimization algorithms},
  author={Schulman, John and Wolski, Filip and Dhariwal, Prafulla and Radford, Alec and Klimov, Oleg},
  journal={arXiv preprint arXiv:1707.06347  }  ,
  year={2017}
}

@article{grpo,
  title={Deepseekmath: Pushing the limits of mathematical reasoning in open language models},
  author={Shao, Zhihong and Wang, Peiyi and Zhu, Qihao and Xu, Runxin and Song, Junxiao and Bi, Xiao and Zhang, Haowei and Zhang, Mingchuan and Li, YK and others},
  journal={arXiv preprint arXiv:2402.03300} ,
  year={2024}
}

@article{reinforce,
  title={Simple statistical gradient-following algorithms for connectionist reinforcement learning},
  author={Williams, Ronald J},
  journal={Machine learning},
  volume={8},
  pages={229--256},
  year={1992},
  publisher={Springer}
}

@misc{reinforce-plus-plus,
      title={REINFORCE++: An Efficient RLHF Algorithm with Robustness to Both Prompt and Reward Models}, 
      author={Jian Hu and Jason Klein Liu and Haotian Xu and Wei Shen},
      year={2025},
      eprint={2501.03262},
      archivePrefix={arXiv},
      primaryClass={cs.CL},
      url={https://arxiv.org/abs/2501.03262}, 
}

@misc{self-dc,
      title={Self-DC: When to Reason and When to Act? Self Divide-and-Conquer for Compositional Unknown Questions}, 
      author={Hongru Wang and Boyang Xue and Baohang Zhou and Tianhua Zhang and Cunxiang Wang and Huimin Wang and Guanhua Chen and Kam-fai Wong},
      year={2025},
      eprint={2402.13514},
      archivePrefix={arXiv},
      primaryClass={cs.CL},
      url={https://arxiv.org/abs/2402.13514}, 
}

@article{adaptive-rag,
  title={Adaptive-rag: Learning to adapt retrieval-augmented large language models through question complexity},
  author={Jeong, Soyeong and Baek, Jinheon and Cho, Sukmin and Hwang, Sung Ju and Park, Jong C},
  journal={arXiv preprint arXiv:2403.14403},
  year={2024}
}

@article{deeprag,
  title={DeepRAG: Thinking to Retrieve Step by Step for Large Language Models},
  author={Guan, Xinyan and Zeng, Jiali and Meng, Fandong and Xin, Chunlei and Lu, Yaojie and Lin, Hongyu and Han, Xianpei and Sun, Le and Zhou, Jie},
  journal={arXiv preprint arXiv:2502.01142},
  year={2025}
}

@article{smartcal,
  title={SMARTCAL: An approach to self-aware tool-use evaluation and calibration},
  author={Shen, Yuanhao and Zhu, Xiaodan and Chen, Lei},
  journal={arXiv preprint arXiv:2412.12151},
  year={2024}
}

@article{smart,
  title={Smart: Self-aware agent for tool overuse mitigation},
  author={Qian, Cheng and Acikgoz, Emre Can and Wang, Hongru and Chen, Xiusi and Sil, Avirup and Hakkani-T{\"u}r, Dilek and Tur, Gokhan and Ji, Heng},
  journal={arXiv preprint arXiv:2502.11435},
  year={2025}
}

@article{ado,
  title={ADO: Automatic Data Optimization for Inputs in LLM Prompts},
  author={Lin, Sam and Hua, Wenyue and Li, Lingyao and Wang, Zhenting and Zhang, Yongfeng},
  journal={arXiv preprint arXiv:2502.11436},
  year={2025}
}

@inproceedings{orpo,
  title={Large language models as optimizers},
  author={Yang, Chengrun and Wang, Xuezhi and Lu, Yifeng and Liu, Hanxiao and Le, Quoc V and Zhou, Denny and Chen, Xinyun},
  booktitle={The Twelfth International Conference on Learning Representations},
  year={2023}
}

@article{protegi,
  title={Automatic prompt optimization with" gradient descent" and beam search},
  author={Pryzant, Reid and Iter, Dan and Li, Jerry and Lee, Yin Tat and Zhu, Chenguang and Zeng, Michael},
  journal={arXiv preprint arXiv:2305.03495},
  year={2023}
}

@article{prompt-agent,
  title={Promptagent: Strategic planning with language models enables expert-level prompt optimization},
  author={Wang, Xinyuan and Li, Chenxi and Wang, Zhen and Bai, Fan and Luo, Haotian and Zhang, Jiayou and Jojic, Nebojsa and Xing, Eric P and Hu, Zhiting},
  journal={arXiv preprint arXiv:2310.16427},
  year={2023}
}

@article{reflexion,
  title={Reflexion: Language agents with verbal reinforcement learning},
  author={Shinn, Noah and Cassano, Federico and Gopinath, Ashwin and Narasimhan, Karthik and Yao, Shunyu},
  journal={Advances in Neural Information Processing Systems},
  volume={36},
  pages={8634--8652},
  year={2023}
}

@article{spo,
  title={Self-supervised prompt optimization},
  author={Xiang, Jinyu and Zhang, Jiayi and Yu, Zhaoyang and Teng, Fengwei and Tu, Jinhao and Liang, Xinbing and Hong, Sirui and Wu, Chenglin and Luo, Yuyu},
  journal={arXiv preprint arXiv:2502.06855},
  year={2025}
}

@article{flashrag,
  author       = {Jiajie Jin and
                  Yutao Zhu and
                  Xinyu Yang and
                  Chenghao Zhang and
                  Zhicheng Dou},
  title        = {FlashRAG: {A} Modular Toolkit for Efficient Retrieval-Augmented Generation
                  Research},
  journal      = {CoRR},
  volume       = {abs/2405.13576},
  year         = {2024}
}

@article{e5,
  author       = {Liang Wang and
                  Nan Yang and
                  Xiaolong Huang and
                  Binxing Jiao and
                  Linjun Yang and
                  Daxin Jiang and
                  Rangan Majumder and
                  Furu Wei},
  title        = {Text Embeddings by Weakly-Supervised Contrastive Pre-training},
  journal      = {CoRR},
  volume       = {abs/2212.03533},
  year         = {2022}
}

@article{verl,
  title={HybridFlow: A Flexible and Efficient RLHF Framework},
  author={Wu, Chuan},
  journal={EuroSys 2025 (30/03/2025-03/04/2025, Rotterdam)},
  year={2025}
}

@inproceedings{kilt,
  author       = {Fabio Petroni and
                  Aleksandra Piktus and
                  Angela Fan and
                  Patrick S. H. Lewis and
                  Majid Yazdani and
                  Nicola De Cao and
                  James Thorne and
                  Yacine Jernite and
                  Vladimir Karpukhin and
                  Jean Maillard and
                  Vassilis Plachouras and
                  Tim Rockt{\"{a}}schel and
                  Sebastian Riedel},
  editor       = {Kristina Toutanova and
                  Anna Rumshisky and
                  Luke Zettlemoyer and
                  Dilek Hakkani{-}T{\"{u}}r and
                  Iz Beltagy and
                  Steven Bethard and
                  Ryan Cotterell and
                  Tanmoy Chakraborty and
                  Yichao Zhou},
  title        = {{KILT:} a Benchmark for Knowledge Intensive Language Tasks},
  booktitle    = {Proceedings of the 2021 Conference of the North American Chapter of
                  the Association for Computational Linguistics: Human Language Technologies,
                  {NAACL-HLT} 2021, Online, June 6-11, 2021},
  pages        = {2523--2544},
  publisher    = {Association for Computational Linguistics},
  year         = {2021},
  url          = {https://doi.org/10.18653/v1/2021.naacl-main.200},
  doi          = {10.18653/V1/2021.NAACL-MAIN.200},
  bibsource    = {dblp computer science bibliography, https://dblp.org}
}

\newpage
\appendix

\end{document}